\documentclass[runningheads]{llncs}

\usepackage{graphicx}

\usepackage{xcolor}
\usepackage{amsmath}
\usepackage{xspace}
\usepackage{amssymb}
\usepackage{pifont}
\usepackage[utf8]{inputenc}
\usepackage{float}

\newcommand*{\eg}{e.g.\@\xspace}
\newcommand*{\ie}{i.e.\@\xspace}

\newcommand{\EE}{\ensuremath{\mathbb{E}}}

\newcommand{\round}[1]{\left(#1\right)}
\newcommand{\brackets}[1]{\left[#1\right]}
\newcommand{\where}{\,\middle|\,}
\newcommand{\angles}[1]{\left<#1\right>}
\newcommand{\abs}[1]{\left|#1\right|}
\newcommand{\floorceil}[1]{\left\lfloor#1\right\rceil}
\newcommand{\cmark}{\ding{51}}%
\newcommand{\xmark}{\ding{55}}%

\newcommand{\epsend}{\epsilon_{\mathit{end}}}

\newcommand{\pos}{p}
\newcommand{\velo}{v}
\newcommand{\acc}{a}
\newcommand{\dx}{\ensuremath{\mathit{v_x}}}
\newcommand{\dy}{\ensuremath{\mathit{v_y}}}
\newcommand{\ax}{\ensuremath{\mathit{a_x}}}
\newcommand{\ay}{\ensuremath{\mathit{a_y}}}
\newcommand{\sigx}{\ensuremath{\sigma_x}}
\newcommand{\sigy}{\ensuremath{\sigma_y}}
\newcommand{\mx}{\ensuremath{m_x}}
\newcommand{\my}{\ensuremath{m_y}}
\newcommand{\dgx}{\mathit{dg}_x}
\newcommand{\dgy}{\mathit{dg}_y}
\newcommand{\dgt}{\mathit{dg}}

\newcommand{\rt}{Racetrack\@\xspace}
\newcommand{\selfplay}{self-play\@\xspace} 

\begin{document}
\title{Tracking the Race Between Deep Reinforcement Learning and Imitation  Learning --\\ Extended Version}%
\titlerunning{Tracking the Race Between DRL and IL}
%
\author{Timo P. Gros
\and
Daniel Höller \and
Jörg Hoffmann \and
Verena Wolf}
\authorrunning{T. Gros et al.}
\titlerunning{Tracking the Race Between Deep RL and Imitation  Learning}
%
\institute{Saarland University, Saarland Informatics Campus, 66123 Saarbrücken, Germany
\email{\{timopgros, hoeller, hoffmann, wolf\}@cs.uni-saarland.de}\\
\url{https://mosi.uni-saarland.de, http://fai.cs.uni-saarland.de}
}
\maketitle              

\begin{abstract}

Learning-based approaches for solving large sequential decision making problems have become popular in recent years. The resulting agents perform differently and 
their characteristics depend on those of the underlying learning approach.

Here, we consider a benchmark planning problem from the reinforcement learning domain, the \rt, to investigate the properties of agents derived from different deep (reinforcement) learning approaches. We compare the performance of deep supervised learning, in particular imitation learning, to reinforcement learning for the \rt model.  

We find that imitation learning yields agents that follow more risky paths.
In contrast, the decisions of deep reinforcement learning  are more foresighted, \ie, avoid states in which fatal decisions are more likely.
Our evaluations show that for this sequential decision making problem, deep reinforcement learning performs best in many aspects even though for imitation learning  optimal decisions are considered.

\end{abstract}

\section{Introduction}
In recent years, deep learning (DL) and especially deep reinforcement learning (DRL) have been applied with great successes to the task of learning near-optimal policies for 
sequential decision making problems. 
DRL has been applied to various applications such as Atari games~\cite{mnih-atari-2013,Mnih2015}, Go and  Chess~\cite{silver:etal:nature-16a,silver:etal:science-18,silver:etal:nature-17}, or Rubic’s cube~\cite{agostinelli-et-al-nmi2019}.
It relies on a feedback loop between self-play and the improvement  of the 
current strategy by reinforcing decisions that lead to  good performance. 

Passive imitation learning (PIL) is another well-known   approach to solve sequential decision making problems, where   a policy is  learned based on training data that is labeled by an expert~\cite{schaal1999imitation}.  
An extension of this approach is  active imitation learning (AIL),
where after an initial phase of passive learning, additional data is iteratively generated by  exploring the 
state space based on the current strategy and subsequent expert labeling  \cite{JMLR:v15:judah14a,ross2011reduction}. 
AIL has successfully been applied to common reinforcement learning benchmarks such as cart-pole or bicycle-balancing~\cite{JMLR:v15:judah14a}.

Sequential decision making problems are typically  described by 
 Markov decision processes (MDPs).
 During the simulation of an MDP,  the set of those  states that will be visited in the future depend on current decisions.
In PIL, the agent, which represents a policy, is trained by iterating over the given expert data set, whose distribution does not generally resemble this dependence. 
AIL extends the data with sequentially generated experiences.
Hence, the  data is more biased towards sequentially taken decisions.  
In contrast, DRL does not rely on expert data at all, but simply alternates between  exploitation of former experiences and exploration.  
It is a priori not obvious which method achieves the best result for a particular sequential decision making problem.
 

Here we aim at an in-depth study of empirical learning agent behavior
for a range of different learning frameworks. Specifically we are
interested in differences due to the sequential nature of action
decisions, inherent in reinforcement learning and active imitation learning but not in passive imitation learning. 
To be able to study and understand algorithm behavior in detail, we conduct our investigation in a simple benchmark problem, namely \rt.

\rt is originally a pen and paper game, adopted as a benchmark in AI
sequential decision making for the evaluation of MDP solution
algorithms
\cite{BartoBS95,Bonet01,DBLP:conf/aips/PinedaZ14,Sutton1998}.
A map with obstacles is given, and a policy for reaching a goal region
from an initial position has to be found.  Decisions for two-dimensional accelerations are taken sequentially, which requires
foresighted planning.
Ignoring traffic, changing weather conditions, fuel consumption, and
technical details, \rt can be considered a simplified model of autonomous driving control~\cite{forteracetrack}. 
\rt is ideally suited for a comparison of different learning
approaches, because not only the performance of different agents but
also their ``driving characteristics'' can be analyzed.  Moreover, for
small maps, expert data describing optimal policies can be obtained.

%
We train different agents for \rt using DRL, PIL, and AIL and study
their characteristics.  We first apply PIL and train agents
represented by linear functions and artificial neural networks.  As
expert labeling, we apply the $A^*$ algorithm to find optimal actions
for states in \rt.  We suggest different variants of data generation
to obtained more appropriate sample distributions.  For AIL, we use
the DAGGER approach~\cite{ross2011reduction} to train agents
represented by neural networks.  We use the same network architecture
when we apply deep reinforcement learning.  More specifically, we
train deep Q-networks~\cite{Mnih2015} to solve the \rt benchmark.
We compare the resulting agents considering three different aspects: the success rate, the quality of the resulting action sequences, and the relative number of optimal and fatal decisions.

%
%
Amongst other things, we find that, \emph{even though it is based on
  optimal training data, imitation learning leads to unsafe policies,
  much more risky than those found by RL.} Upon closer inspection, it
turns out that this apparent contradiction actually has an intuitive
explanation in terms of the nature of the application and the
different learning methods: \emph{to minimize time to goal, optimal
  decisions navigate very closely to dangerous states.} This works
well when taking optimal decisions throughout -- but is brittle to
(and thus fatal in the presence of) even small divergences as are to
be expected from a learned policy. We believe that this characteristic
might carry over to many other applications beyond \rt.


The outline of our paper is the following: We first introduce the \rt
domain (Section~\ref{sec:rt-game}).  Then we introduce the DAGGER
framework and deep Q-learning (Section~\ref{sec:learning}), before we describe our application to
the \rt domain (Section~\ref{sec:training}). In Section~\ref{sec:results}, we present our
experiments and findings. We finally draw a conclusion and present
future work in Section~\ref{sec:conclusion}.

This report is an extended version of the conference paper by Gros et al.~\cite{Gros2020}. 

\section{\rt}
\label{sec:rt-game}
\begin{figure}
	\begin{center}
		\includegraphics[width = \columnwidth]{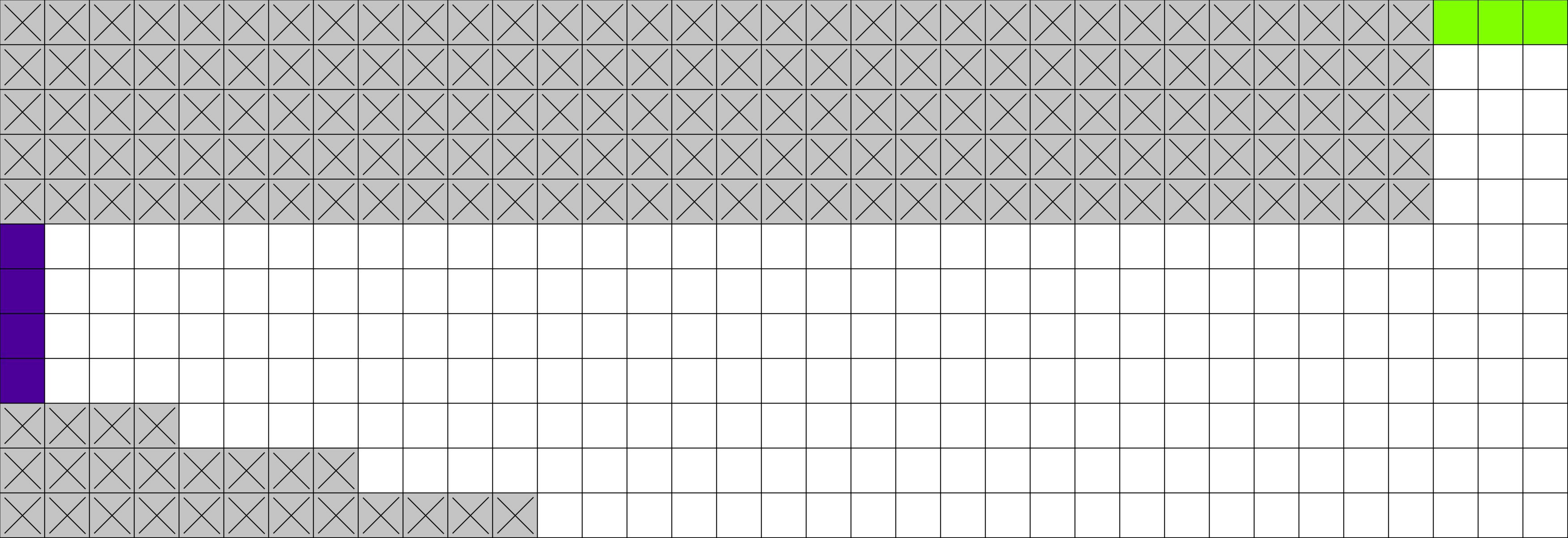}
		\caption{Example of a \rt map: goal line is green,  start line is purple.}
		\label{fig:example_map}
	\end{center}
\end{figure}

\rt has been used as a benchmark in the context of planning~\cite{Bonet01,DBLP:conf/aips/PinedaZ14} and reinforcement learning~\cite{BartoBS95,Sutton1998}.
It can be played on different maps. The example used throughout the paper is displayed in Figure~\ref{fig:example_map}.

\subsection{The \rt Game}

At the beginning of the game, a car is placed randomly at one of the discrete positions on the start line (in purple) with zero velocity. 
In every step it can speed up, hold the velocity or slow down in $x$ and/or $y$ dimension.
Then, the car moves in a straight line with the new velocity from the old position to a new one,
where we discretize the maps into   cells.  
The game is lost when the car crashes, which is the case when either 
(1) the new position itself is a wall position or outside the map, or
(2) the straight line between the old and new position intersects with a wall, \ie the car drives through a wall on its way to the new position. 
The game is won when the car either stops at or drives through the goal line (in green). 

\subsection{Markov Decision Process}\label{sec:MDP}
Given a \rt map, the game can be modeled as a Markov decision process. 
\paragraph{States.} The current state  is uniquely defined by the position $\pos=(x,y)$ and the velocity $\velo=(v_x,v_y)$. 
\paragraph{Actions.} Actions represent the acceleration $\acc$. As the car can be accelerated  with values $\{-1, 0, 1\}$ in the $x$ and in the $y$ dimension, there are exactly $3^2=9$ different actions available in every state. 
\paragraph{Transitions.} We assume a wet road, so with a chance of $0.1$, the acceleration cannot be applied, \ie $\acc = (0,0)$. Otherwise, with probability $0.9$, the acceleration is as selected by the action. The new velocity $\velo' = \round{\dx', \dy'}$ is given by the sum of the acceleration $\acc = (\ax, \ay)$ and the current velocity. %
The new position $\pos' = \round{x', y'}$ is given by adding   $\velo'$ to the current position, \ie
\begin{align*}
	\dx' &= \dx + \ax, && x'  = x + \dx,  \\
	\dy' &= \dy + \ay,  && y'  = y + \dy. 
\end{align*}
To define several properties, we use a discretization of transitions of the MDP similar to the one of Bonet \& Geffner~\cite{Bonet01}.
%
The corresponding driving trajectory is a sequence of visited positions $$T = \angles{\round{x_0,y_0}, \round{x_1,y_1}, \dots, \round{x_{n-1}, y_{n-1}}, \round{x_n,y_n}},$$ such that
\begin{equation*}
  \resizebox{\linewidth}{!}{%
  $T = 
  \begin{cases}
    \angles{\round{x,y}} & \begin{array}{c}\text{if } \dx' = 0 \text{ and } \dy' = 0 \end{array}\text{ (1)} \\[0.25em]  
    \angles{\round{x,y}, \round{x+\sigx,y}, \round{x+2\cdot\sigx,y} \dots, \round{x',y'}}  & \begin{array}{c}\text{if } \dx' \neq 0 \text{ and } \dy' = 0 \end{array}\text{ (2)} \\[0.25em]
	 \angles{\round{x,y}, \round{x,y+\sigy}, \round{x,y+2\cdot\sigy} \dots, \round{x',y'}}  & \begin{array}{c}\text{if } \dx' = 0 \text{ and } \dy' \neq 0 \end{array}\text{ (3)} \\[0.25em]
   \angles{\round{x,y}, \round{x + \sigx,\floorceil{y+\my}}, \round{x + 2\cdot \sigx,\floorceil{y+2\cdot\my}} \dots, \round{x',y'}}  & \begin{array}{c}\text{if } \dx' \neq 0 \text{ and } \dy' \neq 0 \\
   \text{ and } \abs{\dx'} \geq \abs{\dy'} \end{array}\text{ (4)} \\[0.25em] 
  \angles{\round{x,y}, \round{\floorceil{x + \mx},{y+\sigy}}, \round{\floorceil{x + 2\cdot \mx},{y+2\cdot\sigy}} \dots, \round{x',y'}}  & \begin{array}{c}\text{if } \dx' \neq 0 \text{ and } \dy' \neq 0 \\
  \text{ and } \abs{\dx'} < \abs{\dy'} \end{array}\text{ (5),} 
\end{cases}$}
\end{equation*}
\noindent where $\sigx = \mathit{sgn}(\dx')$, $\sigy = \mathit{sgn}(\dy')$ and $\mx = \frac{\dx'}{\abs{\dy'}}$, $\my = \frac{\dy'}{\abs{\dx' }}$~\cite{zenodoArchive}.

If either the vertical or horizontal speed is $0$, exactly all grid coordinates between  $\pos=\round{x,y}$ and $\pos'=\round{x',y'}$ are contained in the trajectory. 
Otherwise, we consider $n$ equidistant points on the linear interpolation between the two positions and for each one round to the closest position on the map. 
While in the original discretization $n = \abs{\dx'}$~\cite{Bonet01}, in this model it is given by $\text{max}\round{\abs{\dx}, \abs{\dy}}$. 
The former is problematic when having a velocity which moves less into the $x$ than into the $y$ direction, as then only few points will be contained in the trajectory and counterintuitive results may be produced. 

We consider a transition to be \emph{valid}, if and only if it does not crash, \ie no position $p \in T$ is either a wall or outside of the map. 
A transition is said to \emph{reach the goal}, if and only if one of the positions $p \in T$ is on the goal line. 
Additionally, a transition cannot be invalid \emph{and} reach the goal. 
If a transition fulfills the conditions for both, only the one that was fulfilled first  holds. 
In words: if a car has already reached the goal, it cannot crash anymore and vice versa. 

A transition leads to a state with   new position $p'$ if it is valid and does not reach the goal. If it is invalid, it leads to a bottom state $\bot$ that has no further transitions. Otherwise, \ie it is reaching the goal, it leads to the goal state $\top$.

\paragraph{Rewards/Costs.}
As we consider both, planning and learning approaches, we define the following two cost functions: 
For planning we consider a uniform cost function, such that an optimal planner will find the shortest path to reach the goal line. %
For reinforcement learning,  we consider 
   a reward function that is positive if the step reaches the goal, negative if the step is invalid and $0$ otherwise. More concretely, we chose 
\begin{align*}
	R\round{s \xrightarrow{\round{\ax,\ay}}s'} = 
	\begin{cases}
		\hfill100 & \text{if } s' = \top \\ 
		\hfill-50 & \text{if } s' = \bot \\
		\hfill0 & \text{otherwise}
	\end{cases}
\end{align*}
for the reward of a transition.

As reinforcement learning makes use of discounting, both functions motivate to reach the goal as fast as possible.

\subsection{Simulation}
\label{sec:rt_simulation}
 For a given map, we consider several variants of the simulation.
 \begin{enumerate}
 	\item Normal start (NS) versus random start (RS): Usually a game starts on the start line, but we also consider the slightly more difficult task of starting on a random (valid) position on the map.
 	\item Zero velocity (ZV) versus random velocity (RV): Usually a game starts with velocity $(0,0)$, but we further use a variant starting with a random velocity $(\dx, \dy)$ between $0$ and a given upper bound.
 	\item Noisy (N) versus deterministic (D): Usually the chosen acceleration is applied with the rules given above. When the deterministic option is set, the chosen acceleration is always applied without assuming a wet road, \ie without a chance to ignore the acceleration and keep the velocity unchanged. 
 \end{enumerate}

\section{Learning -- Approaches}\label{sec:learning}

We consider two different learning approaches that are based on different principles. 
Imitation learning is based on labeled training data, while deep reinforcement learning is based on \selfplay without prior knowledge.

\subsection{Imitation Learning}
We consider both passive and active imitation learning. For passive imitation learning, we use 
(1) logistic regression (LR) and linear discriminant analysis (LDA) to train linear functions, and 
(2) stochastic gradient descent to train neural networks. 
To represent the class of active imitation learning algorithms, we consider DAGGER~\cite{ross2011reduction}. 
\paragraph{DAGGER.} 
\emph{Dataset Aggregation} (DAGGER) is a meta-algorithm for active imitation learning. 
The main idea of DAGGER is to mitigate  the problem related to the disruption of the independently identical distributed (i.i.d.) assumption in passive imitation learning for sequential decision making.
 The trained agent is then used to iteratively sample more labeled data to train a new neural network.
 The algorithm starts with a pre-trained neural network using the following steps:
 
\begin{itemize}
	\setlength{\arraycolsep}{10ex} 
	\setlength{\itemsep}{-8pt} 
	\item[(i)]  It follows the current action policy to explore the state space.\\
	\item[(ii)] For every visited state, it uses an expert to find the action that shall be imitated.\\
	\item[(iii)]  It adds the pairs of state and action to the training set, and\\
	\item[(iv)]  trains a new policy on the enlarged data set. 
\end{itemize}
Step (i) can be varied via a hyper-parameter $\beta \in [0,1]$ that sets the ratio of following the current policy or the expert for exploration. With $\beta = 0$ it follows the current policy only.  
Step (iii) can be done with any thinkable expert and 
step (iv) can be done with any training procedure. 

\subsection{Deep Reinforcement Learning}

While there are many different approaches of deep reinforcement learning, \eg policy-based methods~\cite{mnih2016asynchronous} or methods based
 on Monte Carlo tree search~\cite{silver:etal:nature-16a,silver:etal:nature-17}, we here focus on the value-based approach of  \emph{deep Q-learning}~\cite{Mnih2015}.
 
 \paragraph{Deep Q-learning.} 

Given an MDP, we train an agent which represents a policy such that 
  the expected cumulative reward of the MDP's episodes is maximized. 
As (potentially) a race can last forever, the task is a continuing one~\cite{Sutton1998} and the accumulated future reward, the so-called \emph{return}, of step $t$ is therefore given by 
$
 G_t = \sum\nolimits_{i=t}^{\infty} \gamma^i \cdot R_{i+1},
$
  where $\gamma$ is a discount factor with $\gamma\in [0,1]$ and we assume that $R_{i+1}$ is the reward obtained during the transition
from the state $S_i$ to state $S_{i+1}$ for $i\in\{0,1,\ldots\}$~\cite{Sutton1998}.

For a fixed state $s$, an action $a$, and a policy $\pi$, the  \emph{action-value} $q_\pi(s,a)$ gives  the expected return    that is achieved by taking action $a$ in state $s$ and following the policy $\pi$ afterwards, \ie
$$
\textstyle	q_\pi(s,a) = \EE_\pi\brackets{G_t \where S_t = s, A_t = a} = \EE_\pi\brackets{\sum\nolimits_{k=0}^{\infty} \gamma^k R_{t+k+1} \where S_t = s, A_t = a}.
$$

We write $q_*(s,a)$ for the \emph{optimal action-value} function that maximizes the expected return.
%
The idea of \emph{value-based} reinforcement learning methods is to find an estimate $Q(s,a)$ of the optimal action-value function. 
Artificial neural networks can express complex non-linear relationships and are able to generalize. 
Hence, they have become popular for function approximation.
We estimate the Q-value function  using a neural network with weights $\theta$, a so-called deep Q-network (DQN)~\cite{mnih-atari-2013}.
We denote the DQN by $Q(s,a;\theta)$
 and optimize it w.r.t. the target 
\begin{equation}
 \textstyle	 y(s,a;\theta)=\EE \brackets{R_{t+1} + \gamma \cdot \max_{a'} Q(S_{t+1},a';\theta)\mid S_t = s, A_t = a }.\label{eq:target}
 \end{equation} 
 Thus, in iteration $i$ the corresponding loss function  is 
\begin{equation}  
 \textstyle	 L(\theta_i) = \EE\brackets{\round{ y(S_t,A_t;\theta^-) - Q(S_t,A_t;\theta_i) }^2},
 \label{eq:loss}
\end{equation}
where $\theta^-$ refers to the parameters from some previous iteration,
 with the so-called \emph{fixed target}~\cite{Mnih2015} $y(S_t,A_t;\theta^-)$. 
We optimize the loss function by stochastic gradient descent using an approximation of  $\nabla L(\theta_i)$ \cite{Mnih2015}.


Furthermore, we apply  the idea of \emph{experience replay}~\cite{Mnih2015}.
Instead of directly learning from observations, we store all experience tuples in a data set and sample uniformly from that set. 

We generate our experience tuples by exploring the state space epsilon-greedily, that is, with a chance of $1-\epsilon$ during the Monte Carlo simulation we follow the policy that is implied by the current network weights and otherwise uniformly choose a random action~\cite{Mnih2015}.

In the following, we will use the terms reinforcement learning (RL) and deep reinforcement learning (DRL) interchangeably. 
\label{sec:drl}

\section{Training \rt Agents} \label{sec:training}

In this section we describe the training process of agents based on active and passive imitation learning as well as deep reinforcement learning.



\paragraph{State~Encoding.}
Although a state in the \rt{} problem is uniquely given by the car's position and velocity, we provide several other features that can be used as state encoding to improve the learning procedure.
Instead of giving a complete encoding of the grid to the agent,   the following features will be provided. These features correspond well to the idea of \rt being a model of autonomous driving control.

\begin{itemize}
	\item $d_1$, $\dots$, $d_8$: linear distance to a wall in all directions. These eight distances are distributed equally around the car position 
	and are given analogously to the acceleration, \ie $-1$, $0$ or $1$ in both dimensions.
	\item $\dgx, \dgy$: distance to the nearest goal field in $x$ and $y$ dimension, respectively. 
	\item $\dgt$: total goal distance, \ie $\abs{\dgx} + \abs{\dgy}$.
\end{itemize}
 Together with the position and the velocity, this gives us a total of $15$ features per state. 
We use these features for all considered learning approaches.

\paragraph{Objective Function.}
The learning methods that we consider rely on   two different objective functions: DRL uses  the reward function and imitation learning uses data sets that were optimized w.r.t. the number of steps until the goal is reached.  
As DRL makes use of discounting (see Section~\ref{sec:drl}), the accumulated reward is higher if less steps are taken. Thus, both objective functions serve the same purpose, even though they are not completely equivalent.
 Note that a direct mapping from  the costs used in the planning procedure to the reward structure was not possible.
We tested different reward structures for DRL and found that a negative reward for each single step combined with a positive reward for the goal and a negative reward for invalid states led to 
very poor convergence properties of the training procedure. 
No well-performing alternative was found to the reward structure defined in Section~\ref{sec:MDP} up to scaling.

\subsection{Imitation Learning}

\label{subsec:supervised_learning}
We want to train agents for all simulation scenarios including those where the car starts at an arbitrary position on the map and visits future positions on the map with different velocities. 
Usually, all learning methods are based on the assumption that the data is i.i.d.. 
Data that is generated via simulation of the \rt greatly disrupts this assumption. 
Thus, we propose different approaches for data generation
  to encounter this problem.

\paragraph{Data Sets.}
In the base case, we uniformly sample states and velocities for the simulation scenarios described in Section~\ref{sec:rt_simulation}. The samples are then labeled by an expert. 
This expert basically is a \rt-tailored version of the $A^*$ algorithm to find an optimal action (there might be more than one), \ie acceleration, from the current state.

We further use additional options that can be set when sampling data to address the problem of decisions depending on each other:
\begin{itemize}
		\item Complete trajectory (T): If this option is set, all states on the way to the goal are added to the data set instead of only the current state. 
		\item Exhaustive (E): If the exhaustive option is set, \emph{all} optimal solutions for the specified state are added to the data set. 
		\item Unique (U): Only states having a unique optimal acceleration are added to the data set. 
	\end{itemize}
Option E excludes option T due to runtime constraints as the number of optimal trajectories increases exponentially with the trajectory's length. 

This leads to a total of $6$ different combinations   as displayed in Table~\ref{table:datasetsconfig}.

 \begin{table}[h]
 	\caption{\rt configurations used to create our data sets.}
 	\label{table:datasetsconfig}
 	\begin{center}
 		\begin{tabular}{|c|c|p{7.5 cm}||c|c|c|c|c|}
		\hline
		No &ID& Description & RS & RV & T & E & U \\ \hline \hline
		(1) & RS-RV & Uniform sample from all positions on map and all possible velocities.& \cmark & \cmark & \xmark & \xmark & \xmark \\ \hline

		(2) & NS-ZV-T & Uniform sample from all positions on the start line; combined with zero velocity. All states that were visited on the optimal trajectory to the goal line are included in the data set. & \xmark & \xmark & \cmark & \xmark & \xmark \\ \hline
		(3)& RS-ZV-T & Uniform sample from  all positions on the map; combined with zero velocity. All states that were visited on the optimal trajectory to the goal line are included in the data set.& \cmark & \xmark & \cmark & \xmark & \xmark \\ \hline
		(4) & RS-RV-T & Uniform sample from  all positions on the map and all possible velocities. All states visited on the optimal trajectory to the goal line are included in the data set.& \cmark & \cmark & \cmark & \xmark & \xmark \\ \hline \hline
		(5) & RS-RV-E & Uniform sample from  all positions on the map and all possible velocities. All optimal actions for that state are included in the data set. &\cmark & \cmark & \xmark & \cmark & \xmark \\ \hline 
		(6) & RS-RV-U  & Uniform sample from  all positions on the map and all possible velocities. Only such states that have a unique optimal action are included in the data set. &\cmark & \cmark & \xmark & \xmark & \cmark \\ 
		\hline 
		\end{tabular}
 	\end{center}			
 \end{table}

The first data set contains uniformly sampled (valid) positions and velocities and combines them with a single optimal action. This explores the whole state space equally. The data sets (2) and (3) differ in their starting points. For (2), the car is positioned on the start line, for (3) it might be anywhere on the map. Both sets contain not only the optimal acceleration for this starting state, but for every one  visited on the trajectory from there on to the goal.  To do both, uniformly sample through the state space and take into account the trajectories, (4) starts with a random position and a random velocity but still collects the whole trace. 
The data set (5) includes all optimal solutions instead of just one. Apart from that, (5) is similar to set (1). (6) only includes entries that have a unique optimal next action.  

For each learning method, we train several instances; at least one on each data set. 
Each data set consists of approximately $10^5$ entries. 

\subsubsection{Passive Imitation Learning}

\paragraph{Linear Predictors.}
While deep learning clearly is more powerful than linear learning, linear classifiers have the advantage that their decisions are more transparent. 

We use the package \verb|sklearn| to apply both \emph{Linear Discriminant Analysis} (LDA) and \emph{Logistic Regression} (LR). Together with the six combinations of data sets, this gives $12$ different agents. 

\paragraph{Neural Networks.} \label{sec:training:nn}
We use the \verb|PyTorch| package to train neural networks~\cite{ketkar2017introduction}. We repeatedly iterate over the labeled training data. We use the MSE as loss function. As neural networks tend to overfit when the training iterates over the training data too often, we store the neural network after every iteration. We experimentally found that a maximum iteration number of $20$ is more than sufficient. As we again use every $6$ data sets, this gives us a total of $120$ agents. 

As explained in Section~\ref{sec:rt-game}, a state is represented by $15$ features, which gives us the input size of the network. There are $9$ possible actions. 
As we do not process the game via an image but through predefined features, we only use fully connected layers. 
More sophisticated network structures are only needed for complex inputs such as images. 
For a fair comparison, we  use the same network size for all methods. 
We use two hidden layers of size $64$, resulting in a network structure of $15 \times 64 \times 64 \times 9$.

\subsubsection{Active Imitation Learning}
\paragraph{DAGGER.}

In the case of active imitation learning, we applied
 DAGGER using $\beta = 0$ for all iterations, \ie after the pre-training we followed the trained agent without listening to the expert for exploration. 
To have a fair comparison, DAGGER has the same number of samples as PIL, \ie $10^5$. 
Still, the pre-training is   important for sampling within the first iteration of the algorithm, but the main idea is to generate further entries that are more important for the training of the agent. 
Thus, we pre-trained the agent on each of our data sets and then additionally allowed DAGGER to add $10^5$ samples. 
We split these $10^5$ samples into $20$ iterations. The neural network was trained by performing eight iterations over the data set. 
Our experiments with the networks 
%
%
showed that this is the best trade-off between over- and under-fitting. Again, we store the trained agents after every iteration, giving us a total of $120$ agents for the DAGGER method. 

\subsection{Deep Reinforcement Learning}
\paragraph{Deep Q-learning.}
In contrast to imitation learning, reinforcement learning is not based on data sets and thus is not applied to any of the data sets given in Table~\ref{table:datasetsconfig}. Training is done by \selfplay only; the \rt{} agent chooses its actions using a neural network and applies them to the environment. After every move, the next state (given by the features), a reward as defined in Section~\ref{sec:rt-game} that was achieved for the move, as well as the information whether the episode is terminated are returned to the agent. The agent then uses the loss, defined again by the MSE function, between the accumulated reward and the expected return to correct the weight of the network. 

All imitation learning agents were trained with $10^5$ (new) samples using the same network structure. Therefore, we here restrict the agents to 
(1) $10^5$ entries in the replay buffer, \ie the maximal number of entries an agent can learn from at the same time, and (2) $10^5$ episodes that the agent can play at all.
The neural network is not pre-trained but initiated randomly. 

To have a trade-off between exploration and exploitation, our agent acts $\epsilon$-greedy, \ie with a probability of $\epsilon$ it chooses a random acceleration instead of the best acceleration to explore the state space. As our DRL agent is initiated randomly --~and thus starts without any experience about what good actions are~-- in the beginning of the training phase the focus lies on exploration. We therefore begin our training with $\epsilon = 1$, \ie always choosing a random action. After every episode $i$, we decrease $\epsilon$ exponentially with a factor $\lambda = 0.999$ to shift the focus from exploration to exploitation during training, until a threshold of $\epsilon_{\text{end}} = 0.0001$ is reached, \ie $\epsilon_{i+1} = \max\left(\epsilon_i \cdot \lambda, \epsend\right)$.

To train the agents to not just reach the goal but to minimize the number of steps, we make use of a discount factor $\gamma = 0.99$.

\begin{table}[h]
 	\caption{\rt configurations used to train \rt agents with deep reinforcement learning.}
 	\label{table:trainingconfigs}
	\begin{center}
	\begin{tabular}{|c|c| p{7.5cm}|c|c|}
		\hline
		No &ID & Description & RS &D \\ \hline \hline
		1 & NS-D &Starting on a random position on the start line using the deterministic simulation. & \xmark & \cmark  \\ \hline
		2 & NS-N & Starting on a random position on the start line using the noisy simulation. &\xmark & \xmark  \\ \hline
		3 & RS-D & Starting on a random position on the map  using the deterministic simulation. &\cmark & \cmark   \\ \hline
		4 & RS-N & Starting on a random position on the map  using the noisy simulation & \cmark & \xmark   \\ \hline		 
	\end{tabular}
	\end{center}
\end{table}

Besides the given options of either starting on the start line (NS) or anywhere on the map (RS), DRL can benefit from learning while the noisy (N) version of \rt is simulated instead of the deterministic (D) one. This gives us four different training modes listed in Table~\ref{table:trainingconfigs}.

\begin{figure}[h]
	\begin{center}
			\includegraphics[width = 0.49\columnwidth]{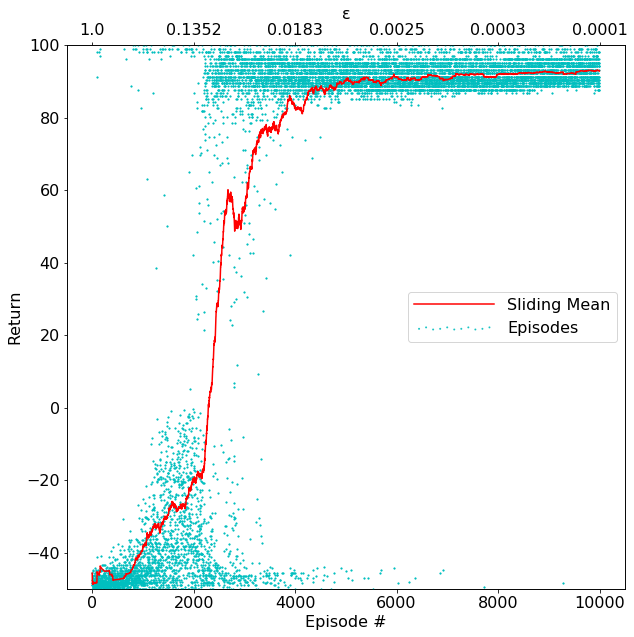} 
			\includegraphics[width = 0.49\columnwidth]{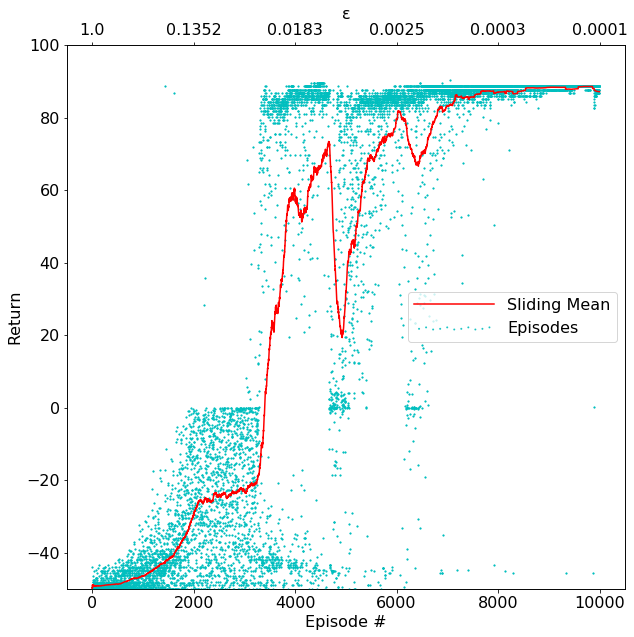}
	\end{center}
	\caption{Training progress of the RL agent. The left graph shows the RS-N mode, the right one displays NS-D. The right plot further displays a temporarily decrease of the return, which is not uncommon during training.} \label{fig:rl_training_progress}
\end{figure}

To determine the best network weights, the average return over the last $100$ episodes during the training process is used. We save the network weights that achieve the best result. 
The training progress is displayed in Figure~\ref{fig:rl_training_progress}. 
Additionally, we save the network weights after running all training episodes, independent from the received average return. This results in total to $8$ different DRL agents. 

 \section{Results} \label{sec:results}

\newlength{\picwidth}
\setlength{\picwidth}{0.44\columnwidth}

For evaluation, we consider all possible combinations given by the simulation parameters as described in Section~\ref{sec:rt_simulation}.
In total, it results in $6$ different simulation settings on which we compare the trained agents. These settings are given in Table~\ref{table:eval}. 
The combinations with  NS and RV are not considered, as they include more starting states where a crash is inevitable than solvable ones.

In the sequel, for each learning method we present the best-performing parameter combination of all those that we tested.
We investigate three aspects of the behavior of the resulting agents: the success rate, the quality of the resulting action sequences, and the relative number of optimal and fatal decisions.

\begin{table}[h!]
	\caption{Configurations on which we evaluate the agents.}
	\begin{center}
	\begin{tabular}{|c|c|p{8cm}||c|c|c|}
		\hline
		No &ID & Description & RS & RV & D \\ \hline \hline
		1 & NS-ZV-D & Starting on a random position on the start line with zero velocity using the deterministic simulation&\xmark & \xmark & \cmark  \\ \hline
		2 & NS-ZV-N &Starting on a random position on the start line with zero velocity using the noisy simulation &\xmark & \xmark &\xmark  \\ \hline
		3 & RS-ZV-D &Starting on a random position on the map with zero velocity using the deterministic simulation &\cmark & \xmark &\cmark  \\ \hline
		4 & RS-ZV-N & Starting on a random position on the map with zero velocity using the noisy simulation &\cmark & \xmark &\xmark   \\ \hline
		5 & RS-RV-D &Starting on a random position on the map with a random velocity using the deterministic simulation &\cmark & \cmark &\cmark   \\ \hline	
		6 & RS-RV-N & Starting on a random position on the map with a random velocity using the noisy simulation&\cmark & \cmark &\xmark   \\ \hline		 
	\end{tabular}
\end{center}
\label{table:eval}
\end{table}

\subsection{Success Rate} 

\begin{figure}[h!]  
		\begin{center}	
		\includegraphics[width = \picwidth]{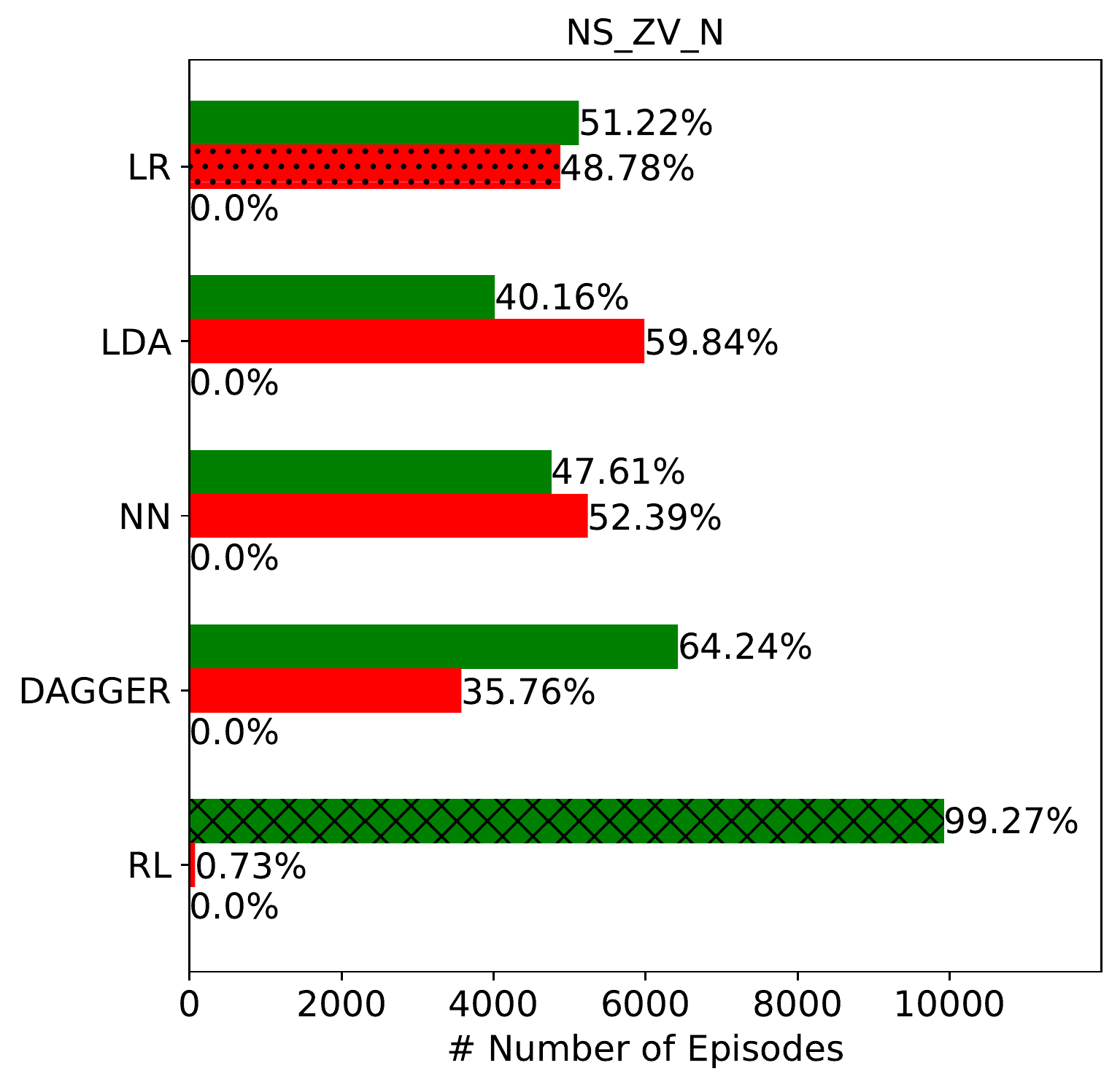}
		\hfill
		\includegraphics[width = 0.44\columnwidth]{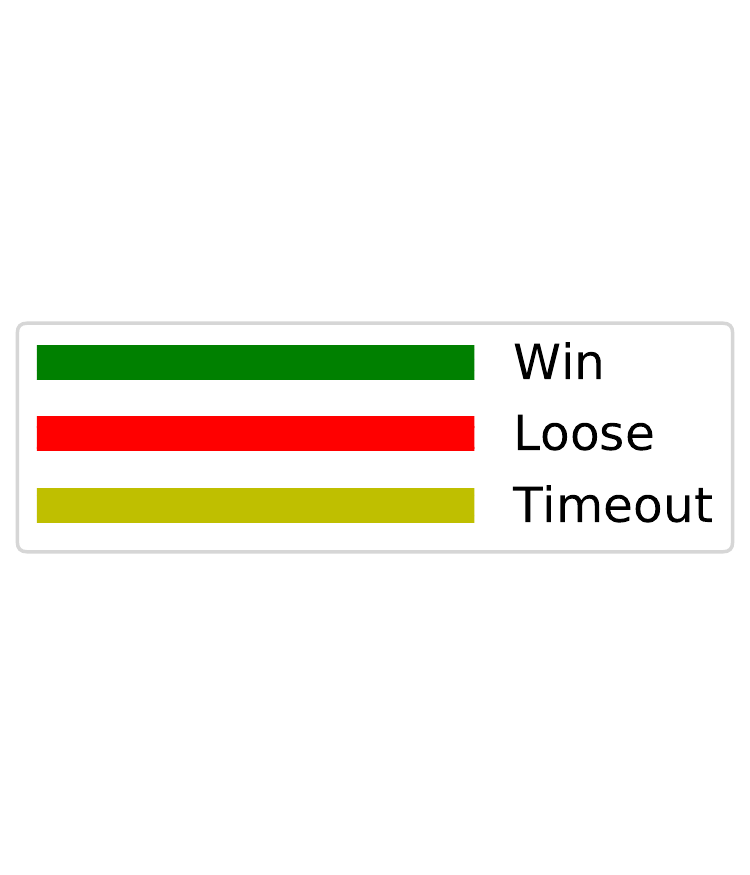}
		\\
		\includegraphics[width = \picwidth]{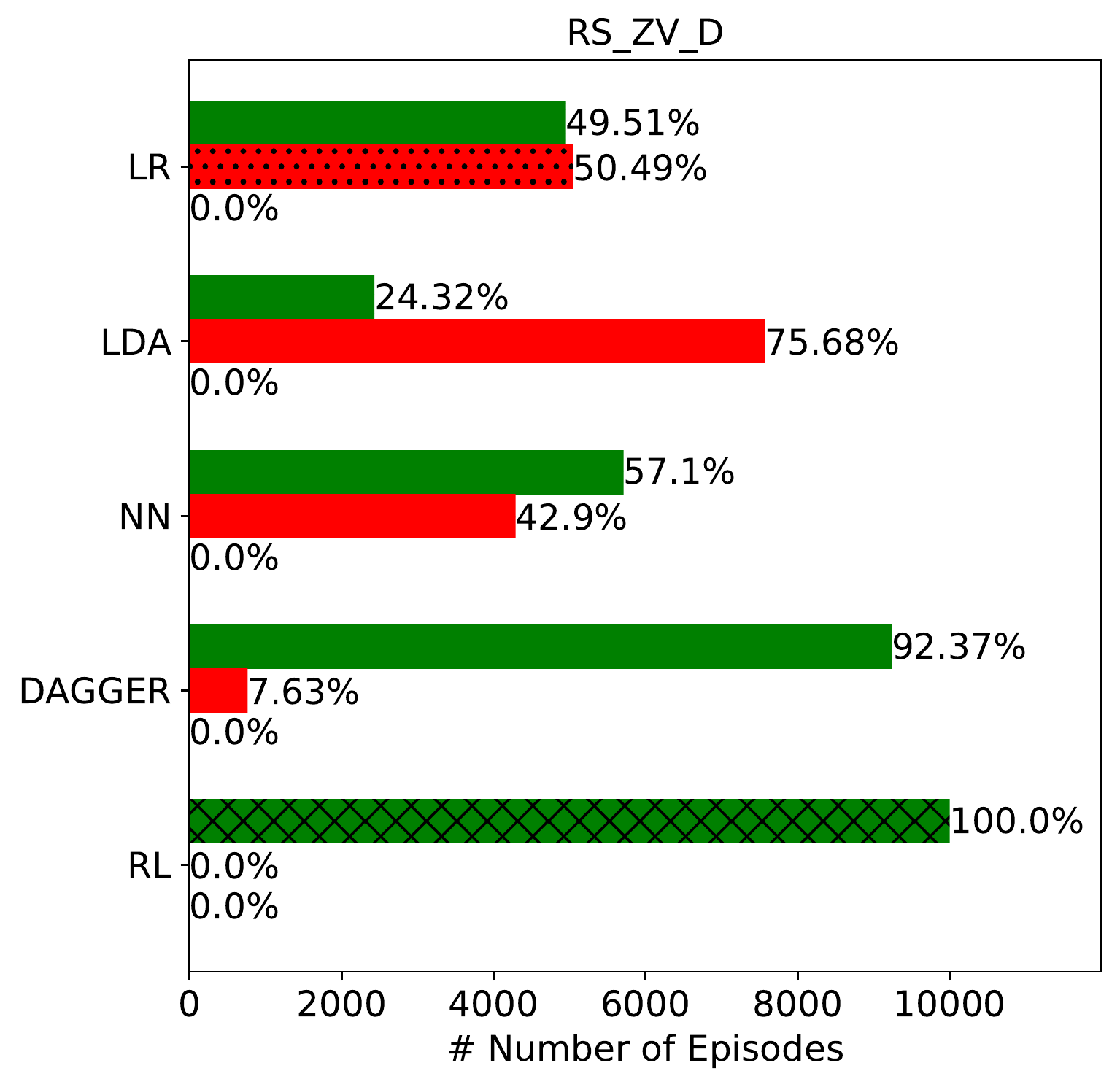}
		\hfill
		\includegraphics[width = \picwidth]{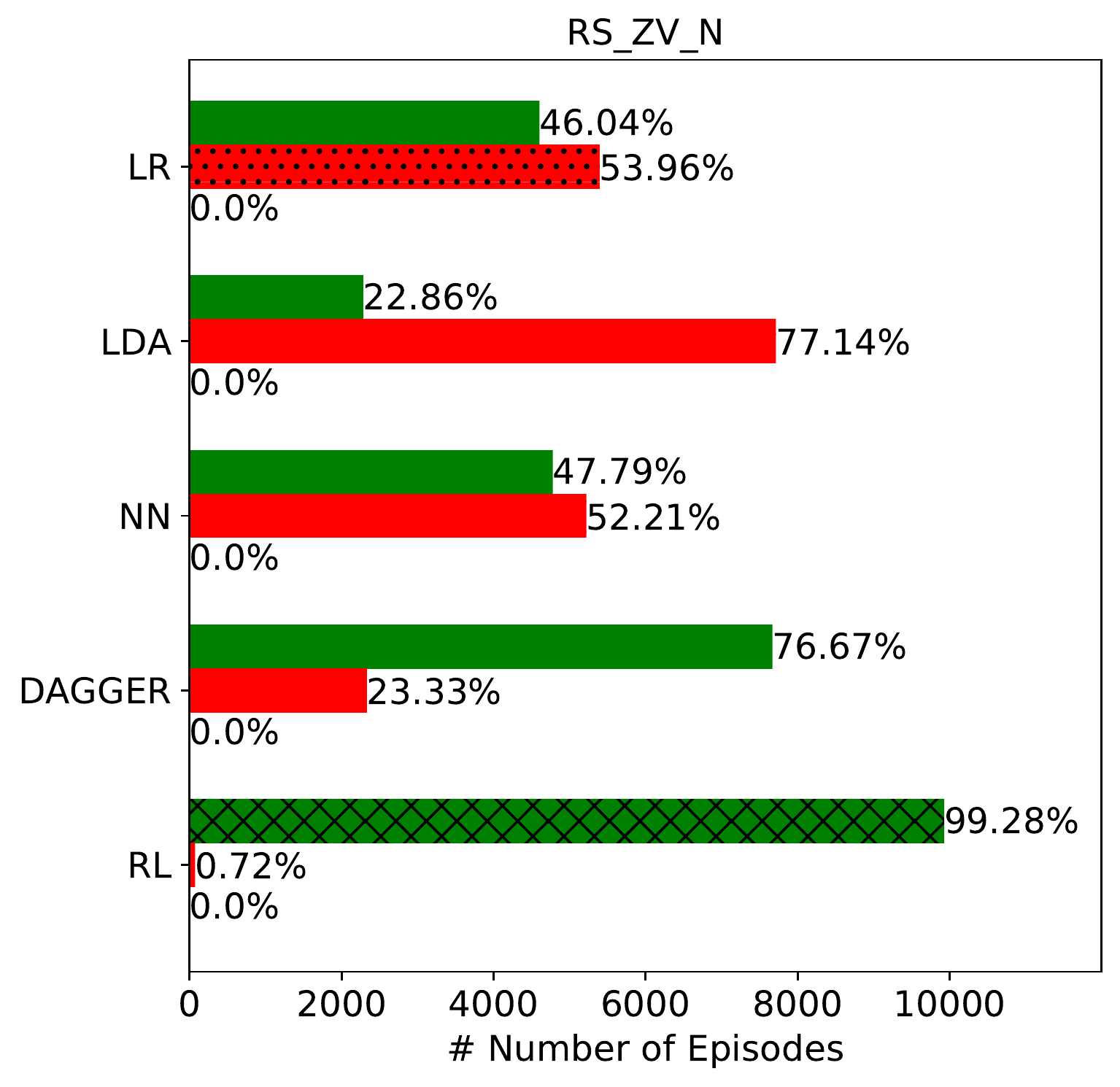}
		\\
		\includegraphics[width = \picwidth]{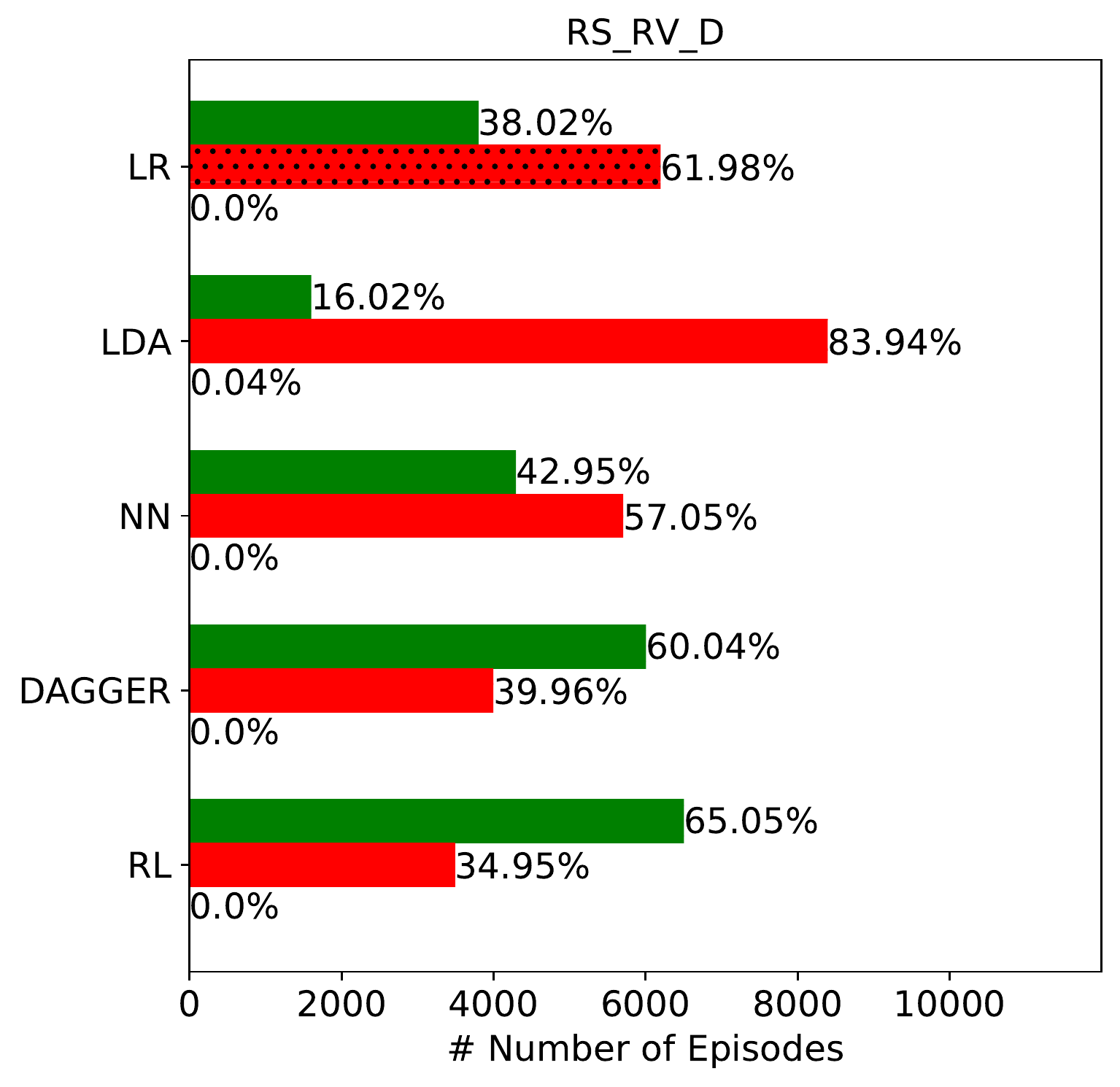}
		\hfill  
		\includegraphics[width = \picwidth]{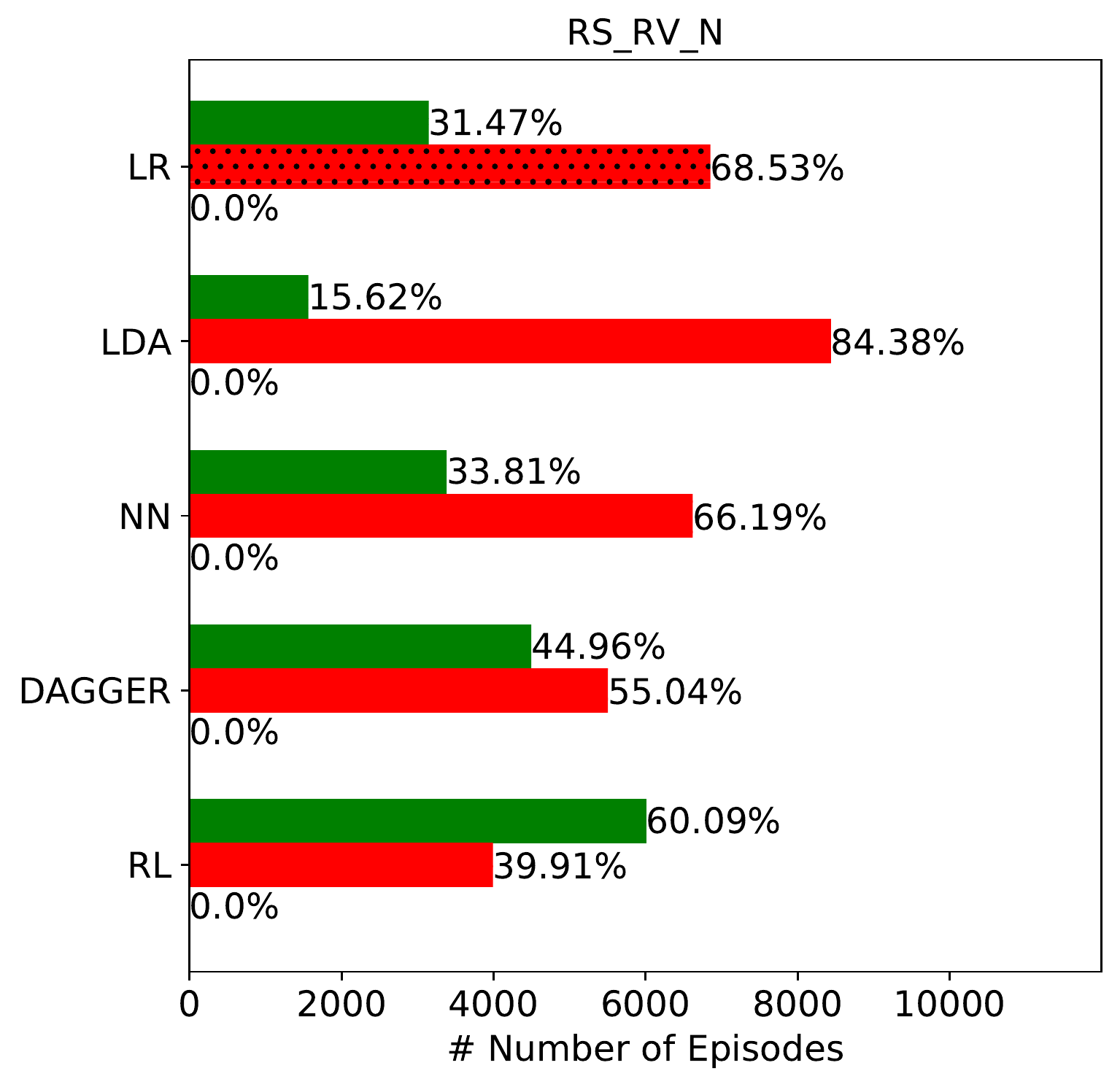}
	\caption{Success rate results for all classes of examined agents.}\label{fig:winloose}    
	\end{center}
\end{figure}  

We first compare how often the agents win a game, \ie reach the goal, or loose, \ie crash into a wall.
We limit the game to $1000$ steps. If an agent then neither succeeded nor failed we count the episode as timed out. 
We compare the agents on $10^4$ simulation runs.
For each single run of the simulation, all agents start in the same initial state.  
The results can be found in Figure~\ref{fig:winloose}.

We omitted the plot for NS-ZV-D, as all of the agents had $100\%$ winning rate. 
The linear agents perform worst. 
Especially with random starting points and velocities, they fail to reach the goal. 
DAGGER outperforms the passive imitation learning agents. This is not surprising, as it has been designed to cope with sequential decision making.

Throughout all settings, the DRL agents perform best. 
They clearly outperform DAGGER, reaching the goal more than $1.5$ times more often in the NS-ZV-N setting.

\subsection{Quality of Action Sequences} 

\begin{figure}[h!]
	\includegraphics[width = 0.49\columnwidth]{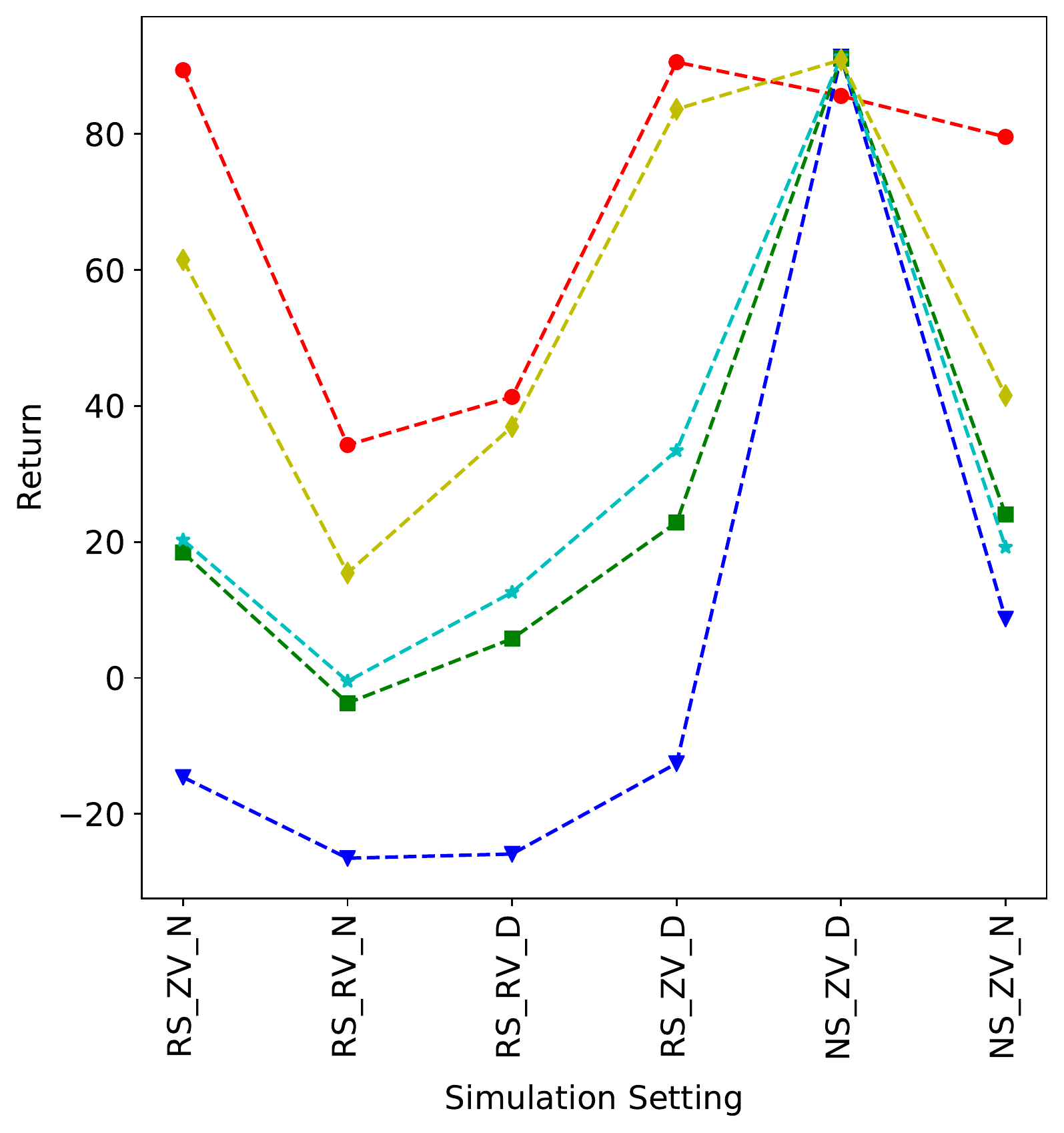}
	\hfill
	\includegraphics[width = 0.49\columnwidth]{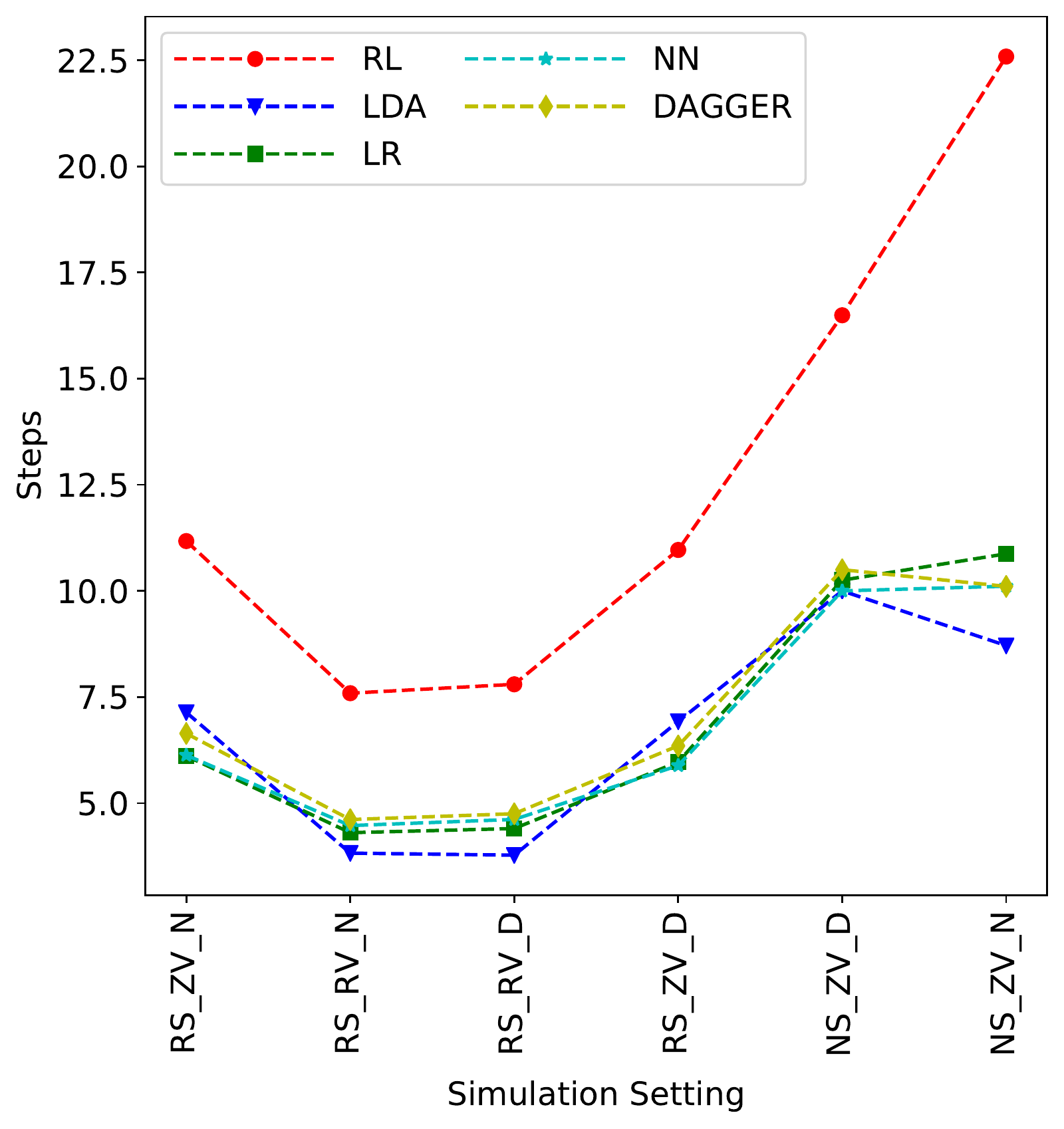}
	\caption{Average reward (left) and average number of needed steps (right) for all classes of agents.}\label{fig:rewardsteps}
\end{figure}

We illustrate results for the quality of the chosen action sequences in Figure~\ref{fig:rewardsteps}.
The left plot gives the  cumulative reward reached by the agents averaged over all runs (also over those that are not successful). DRL clearly achieves the highest cumulative reward.
We remark that the optimal policies computed via A$^*$ give higher cumulative rewards as the goal
is reached faster. However, imitation learning achieves lower results on average as it fails
more often.
 
The right of Figure~\ref{fig:rewardsteps} shows results for the number of steps needed. 
When a car crashes, we are not interested in the number of steps taken.
Therefore -- in this specific analysis -- we only report on successful runs. 
They show that -- while reinforcement learning has the most wins and is the best agent considering the reward objective -- it is consuming the highest number of steps when reaching the goal. 
It even takes more steps than linear classifiers.

\subsection{Quality of Single Action Choices}

   \begin{figure}[t]
 	 \includegraphics[width = 0.49\columnwidth]{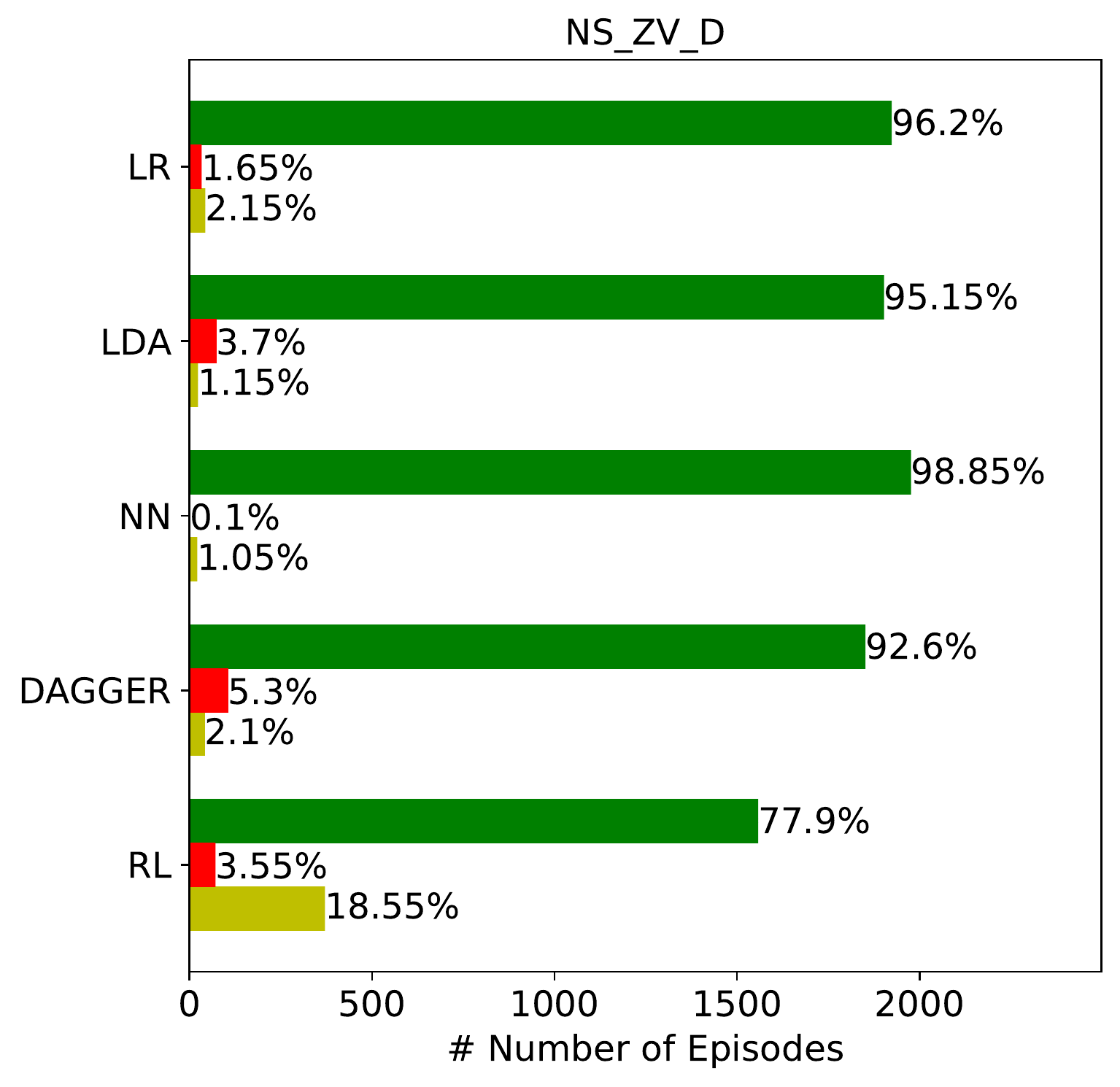}
 	 \hfill
 	 \includegraphics[width = 0.44\columnwidth]{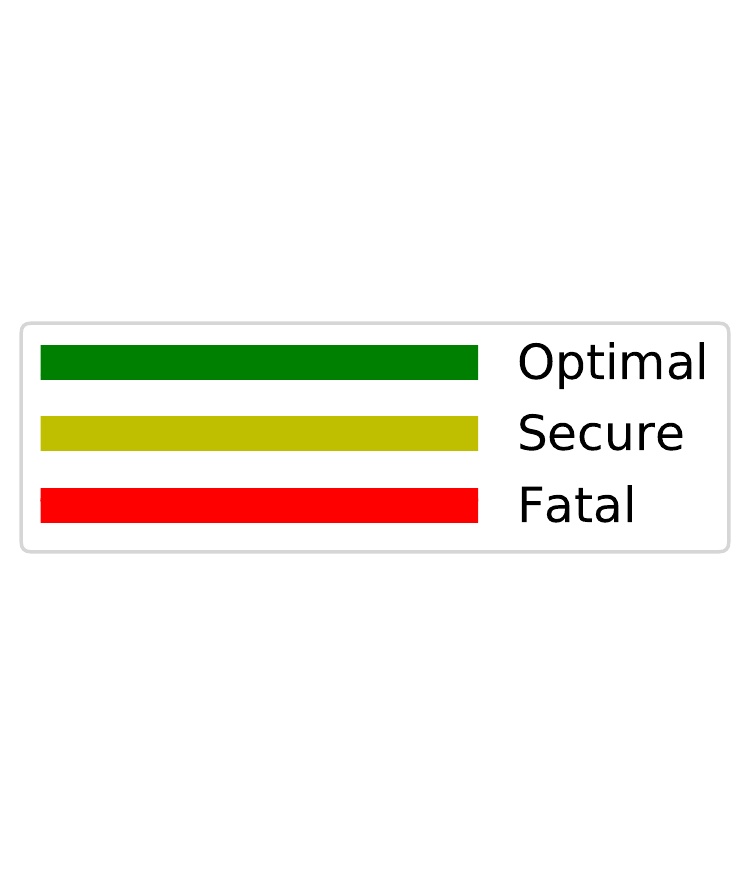}
	\\
 	 \includegraphics[width = 0.49\columnwidth]{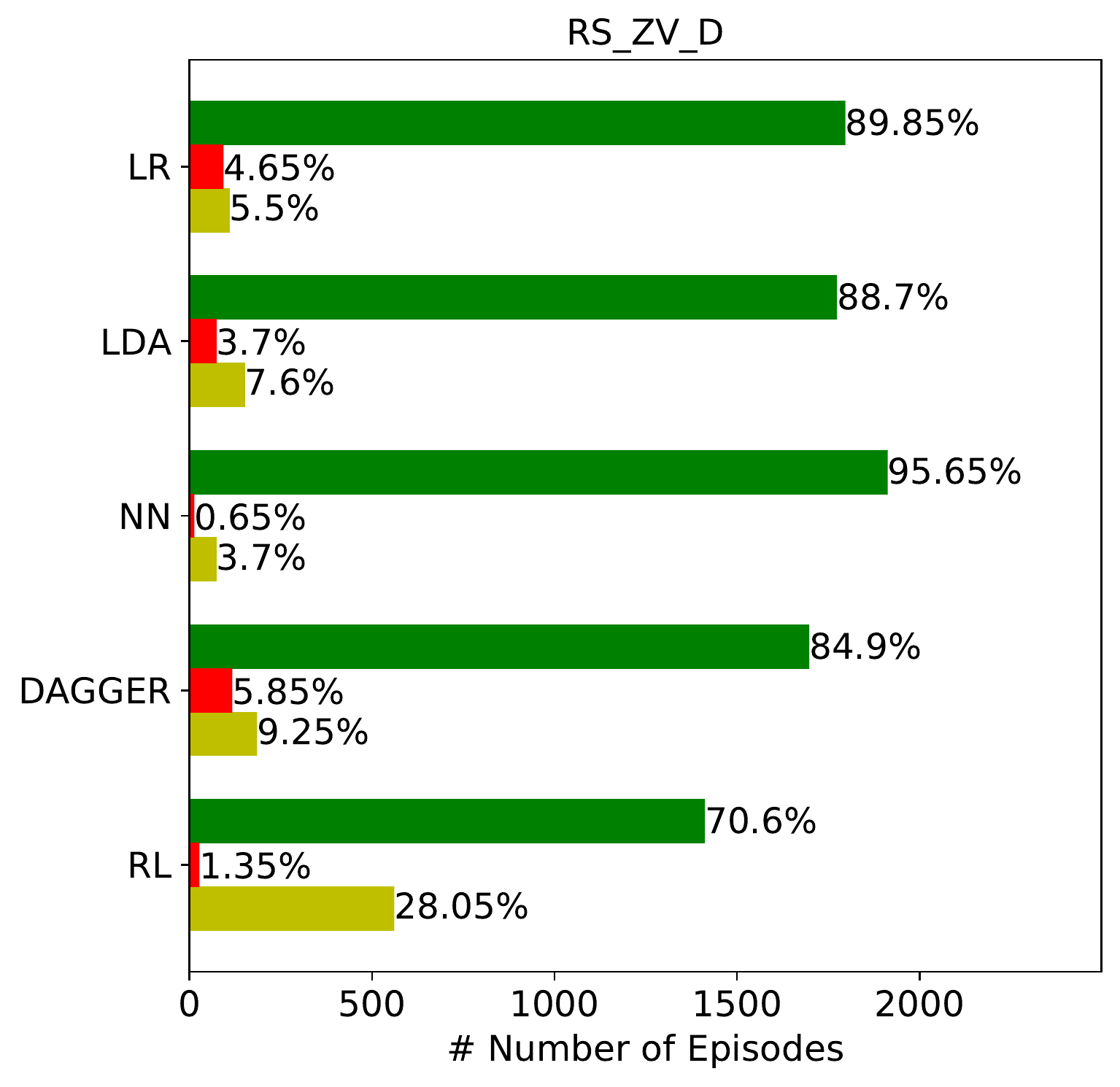}		
 	 \hfill
 	\includegraphics[width = 0.49\columnwidth]{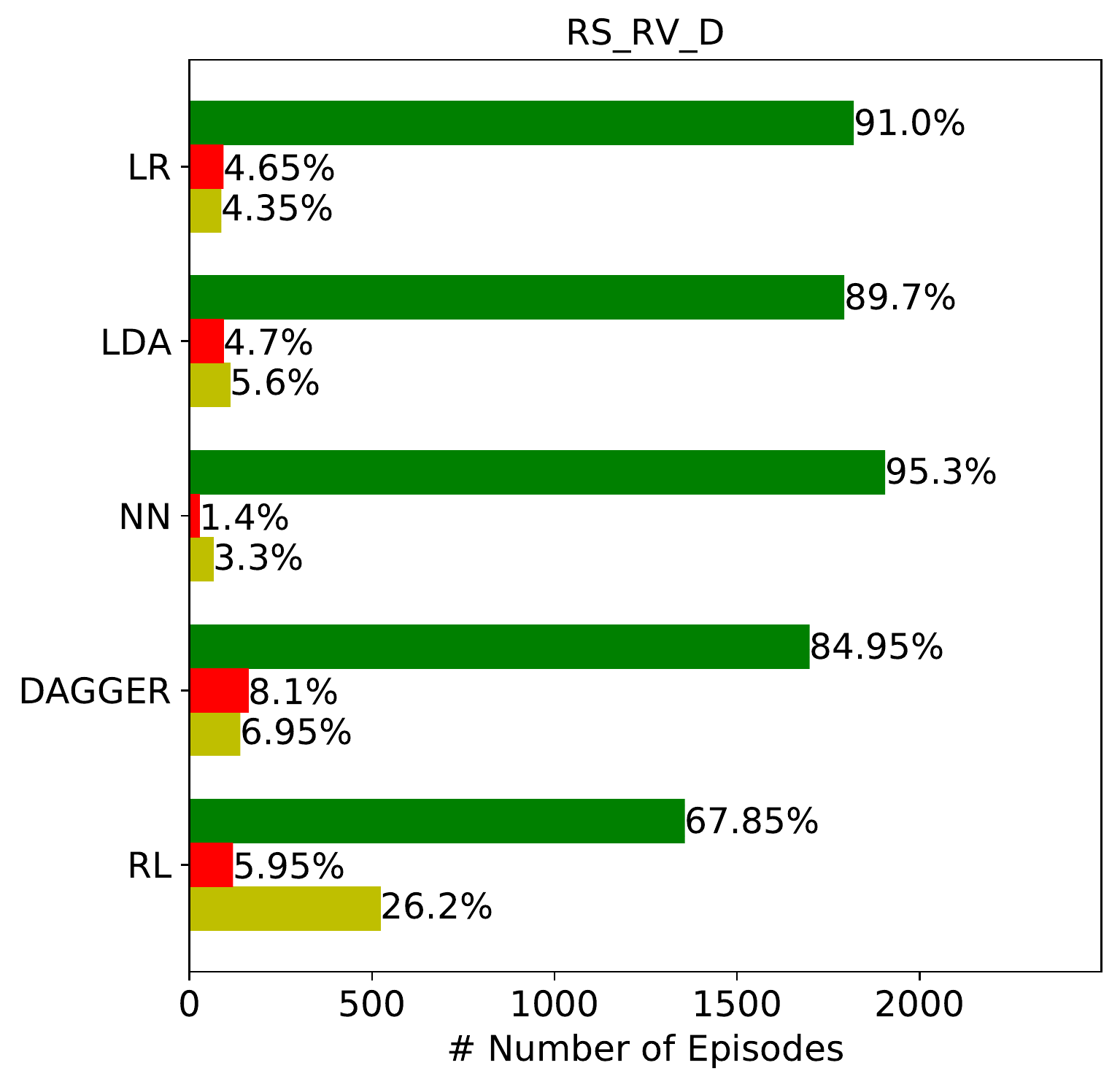}
 	\caption{Quality of selected actions.}\label{fig:actionquality}
 \end{figure}

Next we examine whether the agents choose the optimal acceleration, \ie the acceleration that does not crash and leads to the goal with as few steps as possible,  for different positions and velocities. 
We distinguish between (1) optimal actions, (2) fatal actions that unavoidably lead to a crash, and (3) secure actions that are neither of the former. 
We use the same settings as before, except for the ones with noise, which does not make sense when considering optimal actions, \ie NS-ZV, RS-ZV and RS-RV.

The results are given in Figure~\ref{fig:actionquality}. Especially when we start from a random position on the map, we see that (independent from the setting) passive imitation learning with neural networks selects optimal actions more often than active imitation learning or deep reinforcement learning. 
Interestingly, DAGGER and RL select both secure \emph{and} fatal choices more often than PIL. 

\subsection{Discussion} 

We found that passive imitation learning agents perform poorly (see Figure~\ref{fig:winloose})   even though they select optimal actions most often. 
One reason for this is that the data sets from which they learn contain samples that have not been generated by iteratively improving the current policy.  
Hence, it is not biased towards sequences of dependent decisions leading to good performance.
We have observed that DAGGER and in particular DRL sometimes do not select optimal actions, but those with lower risk of hitting a wall. 
As a result, they need more steps than other approaches before reaching the goal, but the trajectories they use are more secure and they crash less often. 
This is an interesting insight, as all approaches (including PIL) try to optimize the same objective: reach the goal as soon as possible without hitting a wall.   

The fact that both, DAGGER and RL have a relatively high number of fatal actions, but not an increased number of losses, leads us to the assumption that these agents avoid states where they might make fatal decisions, even though these states could help reaching the goal faster. 

\begin{figure}[h!]
	\includegraphics[width = \columnwidth]{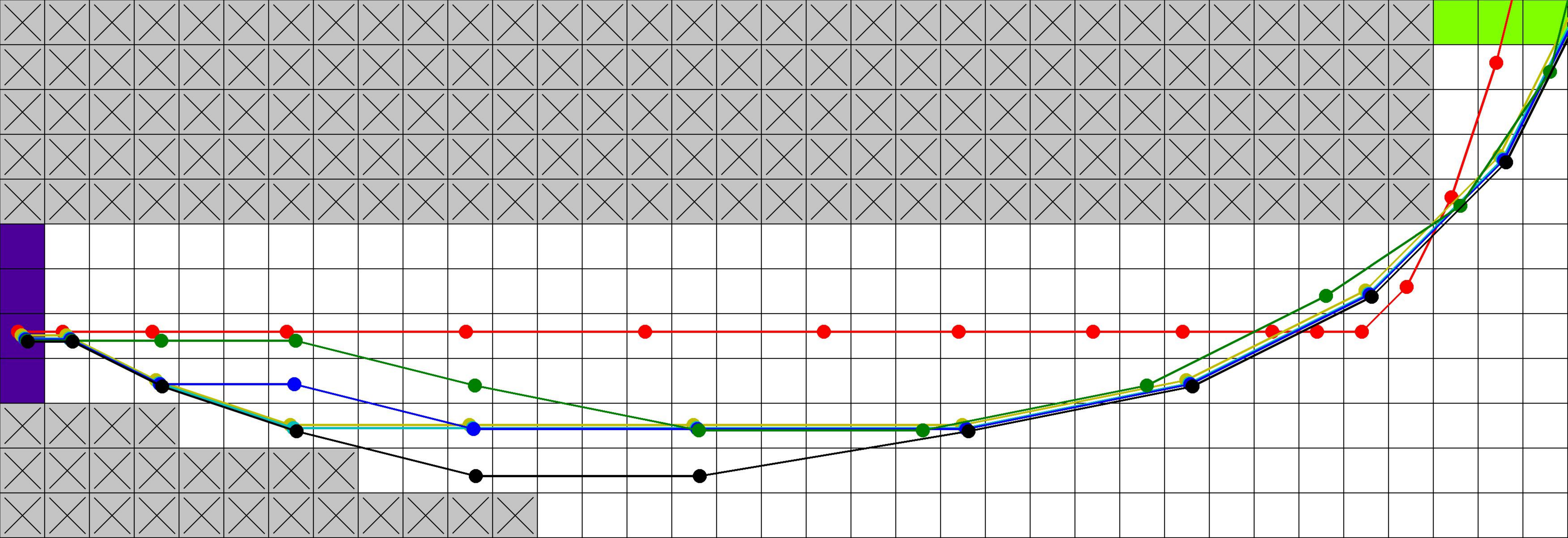}
	\includegraphics[width = \columnwidth]{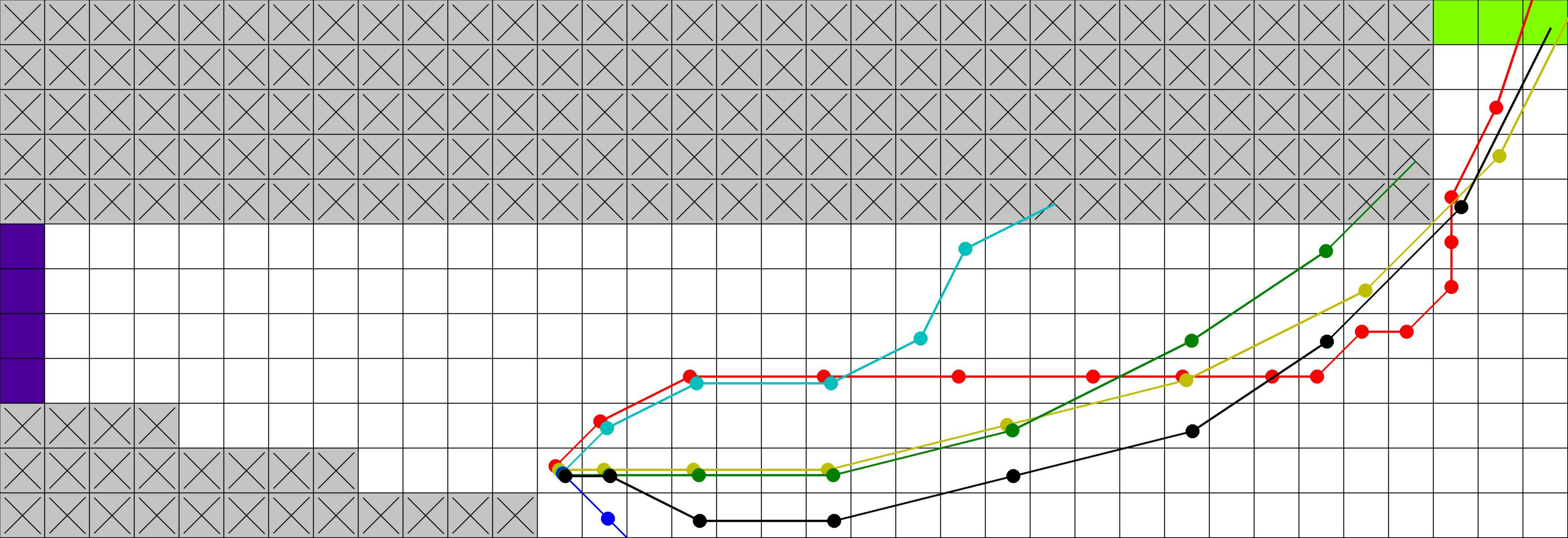}
	\caption{Traces of different \rt agents. The black trajectories are optimal. The other colors are chosen as in Figure~\ref{fig:rewardsteps}. The upper plot is a NS-ZV-D simulation, while the lower shows RS-ZV-D.}
	\label{fig:traces}
\end{figure}

  Figure~\ref{fig:traces} illustrates the paths taken by the different agents  for the easiest case (NS-ZV-D) where all policies reach their goal. DRL differs the most from the optimal (black) trajectory, which describes one of the shortest paths to the goal and obtains the maximum cumulative reward. 
  For the harder setting where a starting point is chosen randomly (RS-ZV-D), only DAGGER and DRL make it to the goal, with DRL using significantly more steps than the optimal agent.

In summary, DRL performs surprisingly well. In some aspects, it performs even better than active imitation learning, which is not only considered a state of the art for sequential decision making~\cite{JMLR:v15:judah14a}, but -- in contrast to DRL -- even has the chance to benefit from expert knowledge. 
\section{Conclusion}\label{sec:conclusion}

We have presented an extensive comparison between different learning
approaches to solve the \rt benchmark.  Even though we provided
optimal decisions during imitation learning, the agents based on deep
reinforcement learning outperform those of imitation learning in many
aspects.

We believe that our observations carry over to other applications, in
particular to more complex autonomous vehicle control algorithms. We
plan to consider extensions of the \rt problem, which include further
real-world characteristics of autonomous driving. We believe that, to
address the difficulties we observed with imitation learning, further
investigations into the combination of expert data sets
and reinforcement learning agents are necessary. 

 Additionally, other methods of guiding the agents to more promising
 solutions during training will be examined, such as reward
 shaping~\cite{laud2004theory}, and their influence on the
 characteristics of the final agent.  Another interesting question for
 future work is whether multi-objective reinforcement learning can be
 used to adjust the agents' behavior in a fine-grained manner.
 
 \section*{Acknowledgements}

This work has been partially funded by DFG grant 389792660 as part of TRR~248 (see \url{https://perspicuous-computing.science}) 

%
%
%
 \bibliographystyle{splncs04}
 \bibliography{bibliography}

\begin{thebibliography}{10}
\providecommand{\url}[1]{\texttt{#1}}
\providecommand{\urlprefix}{URL }
\providecommand{\doi}[1]{https://doi.org/#1}

\bibitem{agostinelli-et-al-nmi2019}
Agostinelli, F., McAleer, S., Shmakov, A., Baldi, P.: Solving the {R}ubik's
  {C}ube with deep reinforcement learning and search. Nature Machine
  Intelligence  \textbf{1}(8),  356--363 (2019)

\bibitem{BartoBS95}
Barto, A.G., Bradtke, S.J., Singh, S.P.: Learning to act using real-time
  dynamic programming. Artificial Intelligence  \textbf{72}(1-2),  81--138
  (1995)

\bibitem{Bonet01}
Bonet, B., Geffner, H.: {GPT}: A tool for planning with uncertainty and partial
  information. In: Proceedings of the IJCAI Workshop on Planning with
  Uncertainty and Incomplete Information. pp. 82--87 (2001)

\bibitem{forteracetrack}
Gros, T.P., Hermanns, H., Hoffmann, J., Klauck, M., Steinmetz, M.: Deep
  statistical model checking. In: Proceedings of the 40th International
  Conference on Formal Techniques for Distributed Objects, Components, and
  Systems ({FORTE}). pp. 96--114. Springer (2020)

\bibitem{zenodoArchive}
Gros, T.P., Hermanns, H., Hoffmann, J., Klauck, M., Steinmetz, M.: {Models and
  Infrastructure used in "Deep Statistical Model Checking"}.
  \url{http://doi.org/10.5281/zenodo.3760098} (2020)

\bibitem{Gros2020}
Gros, T.P., H\"oller, D., Hoffmann, J., Wolf, V.: Tracking the race between
  deep reinforcement learning and imitation learning. In: Proceedings of the
  17th International Conference on Quantitative Evaluation of {SysTems}
  ({QEST}). Springer (2020)

\bibitem{JMLR:v15:judah14a}
Judah, K., Fern, A.P., Dietterich, T.G., Tadepalli, P.: Active imitation
  learning: Formal and practical reductions to i.i.d. learning. Journal of
  Machine Learning Research  \textbf{15}(120),  4105--4143 (2014)

\bibitem{ketkar2017introduction}
Ketkar, N.: Introduction to pytorch. In: Deep learning with {P}ython, pp.
  195--208. Springer (2017)

\bibitem{laud2004theory}
Laud, A.D.: Theory and application of reward shaping in reinforcement learning.
  Tech. rep. (2004)

\bibitem{mnih2016asynchronous}
Mnih, V., Badia, A.P., Mirza, M., Graves, A., Lillicrap, T., Harley, T.,
  Silver, D., Kavukcuoglu, K.: Asynchronous methods for deep reinforcement
  learning. In: International conference on machine learning. pp. 1928--1937
  (2016)

\bibitem{mnih-atari-2013}
Mnih, V., Kavukcuoglu, K., Silver, D., Graves, A., Antonoglou, I., Wierstra,
  D., Riedmiller, M.: Playing {Atari} with deep reinforcement learning. In:
  NIPS Deep Learning Workshop (2013)

\bibitem{Mnih2015}
Mnih, V., Kavukcuoglu, K., Silver, D., Rusu, A.A., Veness, J., Bellemare, M.G.,
  Graves, A., Riedmiller, M., Fidjeland, A.K., Ostrovski, G., Petersen, S.,
  Beattie, C., Sadik, A., Antonoglou, I., King, H., Kumaran, D., Wierstra, D.,
  Legg, S., Hassabis, D.: Human-level control through deep reinforcement
  learning. Nature  \textbf{518}(7540),  529--533 (2015)

\bibitem{DBLP:conf/aips/PinedaZ14}
Pineda, L.E., Zilberstein, S.: Planning under uncertainty using reduced models:
  Revisiting determinization. In: Proceedings of the 24th International
  Conference on Automated Planning and Scheduling ({ICAPS}). pp. 217--225.
  {AAAI} Press (2014)

\bibitem{ross2011reduction}
Ross, S., Gordon, G.J., Bagnell, D.: A reduction of imitation learning and
  structured prediction to no-regret online learning. In: Proceedings of the
  14th International Conference on Artificial Intelligence and Statistics
  ({AISTATS}). {JMLR} Proceedings, vol.~15, pp. 627--635. JMLR.org (2011)

\bibitem{schaal1999imitation}
Schaal, S.: Is imitation learning the route to humanoid robots? Trends in
  cognitive sciences  \textbf{3}(6),  233--242 (1999)

\bibitem{silver:etal:nature-16a}
Silver, D., Huang, A., Maddison, C.J., Guez, A., Sifre, L., van~den Driessche,
  G., Schrittwieser, J., Antonoglou, I., Panneershelvam, V., Lanctot, M.,
  Dieleman, S., Grewe, D., Nham, J., Kalchbrenner, N., Sutskever, I.,
  Lillicrap, T., Leach, M., Kavukcuoglu, K., Graepel, T., Hassabis, D.:
  Mastering the game of {Go} with deep neural networks and tree search. Nature
  \textbf{529},  484--503 (2016)

\bibitem{silver:etal:science-18}
Silver, D., Hubert, T., Schrittwieser, J., Antonoglou, I., Lai, M., Guez, A.,
  Lanctot, M., Sifre, L., Kumaran, D., Graepel, T., Lillicrap, T., Simonyan,
  K., Hassabis, D.: A general reinforcement learning algorithm that masters
  {C}hess, {S}hogi, and {G}o through self-play. Science  \textbf{362}(6419),
  1140--1144 (2018)

\bibitem{silver:etal:nature-17}
Silver, D., Schrittwieser, J., Simonyan, K., Antonoglou, I., Huang, A., Guez,
  A., Hubert, T., Baker, L., Lai, M., Bolton, A., Chen, Y., Lillicrap, T., Hui,
  F., Sifre, L., van~den Driessche, G., Graepel, T., Hassabis, D.: Mastering
  the game of {Go} without human knowledge. Nature  \textbf{550},  354--359
  (2017)

\bibitem{Sutton1998}
Sutton, R.S., Barto, A.G.: Reinforcement Learning: An Introduction. Adaptive
  computation and machine learning, The {MIT} Press, second edn. (2018)

\end{thebibliography}

\end{document}